\documentclass[letterpaper]{article} % DO NOT CHANGE THIS

\usepackage{aaai24}  % DO NOT CHANGE THIS

\usepackage{times}  % DO NOT CHANGE THIS
\usepackage{helvet}  % DO NOT CHANGE THIS
\usepackage{courier}  % DO NOT CHANGE THIS
\usepackage[hyphens]{url}  % DO NOT CHANGE THIS
\usepackage{graphicx} % DO NOT CHANGE THIS

\usepackage{natbib}  % DO NOT CHANGE THIS AND DO NOT ADD ANY OPTIONS TO IT
\usepackage{caption} % DO NOT CHANGE THIS AND DO NOT ADD ANY OPTIONS TO IT
\usepackage{algorithm}
\usepackage{algorithmic}
\usepackage{amsmath}
\usepackage{multirow}
\usepackage{multicol}
\usepackage{newfloat}
\usepackage{listings}
\DeclareCaptionStyle{ruled}{labelfont=normalfont,labelsep=colon,strut=off} % DO NOT CHANGE THIS
\floatstyle{ruled}
\newfloat{listing}{tb}{lst}{}
\floatname{listing}{Listing}
\usepackage{booktabs} % 三线表
\usepackage{makecell} % 表格内换行
\usepackage{xspace}

\newcommand{\modelname}{CCA\xspace}
\newcommand{\fb}{FB15K-237\xspace}
\newcommand{\wn}{WN18RR\xspace}

\title{Knowledge Graph Error Detection with Contrastive Confidence Adaption}
\author{
    Xiangyu Liu\textsuperscript{\rm 1},
    Yang Liu\textsuperscript{\rm 1}, 
    Wei Hu\textsuperscript{\rm 1,2,}\thanks{Corresponding author}
}
\affiliations{
    \textsuperscript{\rm 1} State Key Laboratory for Novel Software Technology, Nanjing University, China\\
    \textsuperscript{\rm 2} National Institute of Healthcare Data Science, Nanjing University, China\\
    \{xyl, yliu20\}.nju@gmail.com, whu@nju.edu.cn
}

\begin{document}

\maketitle

\begin{abstract}
Knowledge graphs (KGs) often contain various errors. 
Previous works on detecting errors in KGs mainly rely on triplet embedding from graph structure. 
We conduct an empirical study and find that these works struggle to discriminate noise from semantically-similar correct triplets.
In this paper, we propose a KG error detection model \modelname to integrate both textual and graph structural information from triplet reconstruction for better distinguishing semantics. 
We design interactive contrastive learning to capture the differences between textual and structural patterns.
Furthermore, we construct realistic datasets with semantically-similar noise and adversarial noise. 
Experimental results demonstrate that \modelname outperforms state-of-the-art baselines, especially in detecting semantically-similar noise and adversarial noise.
\end{abstract}

%====================%
\section{Introduction}

A knowledge graph (KG) is composed of triplets in the form of $(head\ entity, relation, tail\ entity)$, which finds extensive applications in downstream tasks like question answering \cite{EmbedKGQA} and recommender systems \cite{KGRec}.
Existing KGs such as NELL \cite{nell} and Knowledge Vault \cite{knowledge_vault} continuously extract triplets in an automatic way, which inevitably introduces noise.
Detecting these errors holds the potential to improve the quality of KGs.
  
Existing works on KG error detection can be classified into embedding-based and path-based models.
The former \cite{TransE,DistMult,Complex} learns confidence scores based on the representations of entities and relations.
The latter \cite{PTransE,KGTtm} uses paths between entities to evaluate the confidence of triplets.
Different from the task of link prediction \cite{hitter} or triplet classification \cite{KGbert}, error detection focuses on detecting the error triplets in the whole unsupervised KG, aiming to capture the variance of triplets and give an accurate estimate of their confidence.

Current KG error detection models face a significant challenge due to the unavailability of noise patterns and the difficulty in acquiring accurately labeled noise samples for robust supervision.
Negative sampling by replacing entities is widely used in previous works, especially the embedding-based models.
However, real-world scenarios often introduce confusing noise semantically related to correct samples.
Let us see two real error triplets in the \fb dataset \cite{FB15K}: \textit{(George Lopez, profession, Disc jockey)} and \textit{(Majel Barrett, profession, Writer)}.
In the former, the correct tail entity should be `Comedian', and in the latter, the correct tail entity should be `Actress'.
Notably, `Disc jockey' and `Writer' both represent professions, mirroring common human errors.
This form of noise is harder to differentiate and aligns closely with human error tendencies.

\begin{table}
\centering
\small
{
\begin{tabular}{lc|cccc}
\toprule
\multirow{2}{*}{Models} & \multirow{2}{*}{\makecell[c]{Noise\\types}} & \multicolumn{2}{c}{\fb} & \multicolumn{2}{c}{\wn} \\
\cmidrule(lr){3-4} \cmidrule(lr){5-6} & & $K$=1\% & $K$=5\% & $K$=1\% & $K$=5\% \\  
\midrule 
\multirow{2}{*}{CAGED}  & Random  & 0.945 & 0.758 & 0.795 & 0.486 \\
                        & Similar & 0.633 & 0.367 & 0.657 & 0.384 \\
\midrule 
\multirow{2}{*}{TransE} & Random  & 0.904 & 0.726 & 0.630 & 0.434 \\
                        & Similar & 0.611 & 0.303 & 0.503 & 0.361 \\ 
\bottomrule
\end{tabular}}
\caption{Results of our empirical study. 
Error triplets are divided into the ``random'' and ``similar'' groups, based on the methods of replacing entities. 
We show the top-$K$ precision of two typical models on \fb and \wn.}
\label{tab:empirical}
\end{table}

\begin{figure*}  
\centering  
\includegraphics[width=\textwidth]{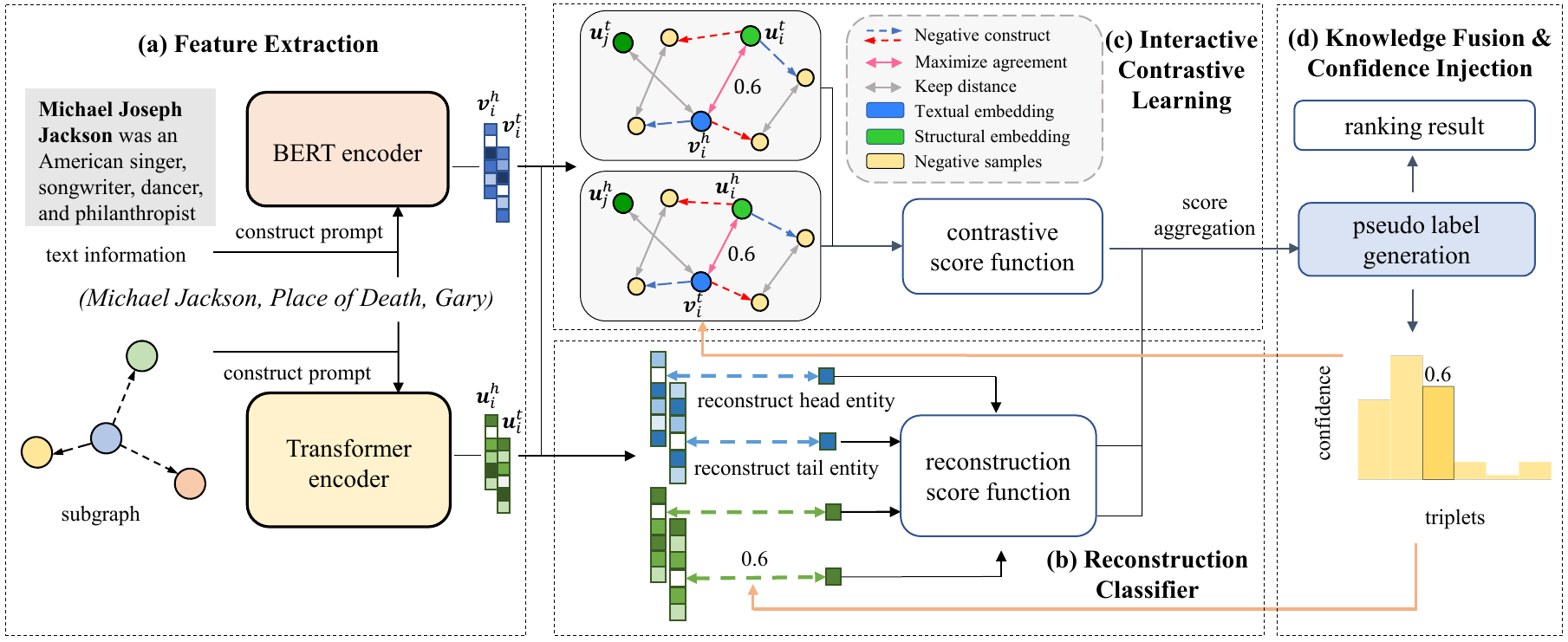}  
\caption{An overview of the proposed model \modelname.
(a) BERT and Transformer-based graph encoders extract textual and graph structural information, respectively.
(b) The reconstruction module classifies error triplets by reconstructing head and tail entities in textual and structural embedding.
(c) Interactive contrastive learning aligns the projection of textual and structural embeddings and recognizes errors by inter-model difference.
(d) The knowledge fusion module takes pseudo labels generated from aggregated results as triplet confidence, which is further injected into the training process. }  
\label{fig:picture001}  
\end{figure*}

We conduct a further empirical study to explore the performance of existing works on two specific types of noise: random noise and semantically-similar noise.
Random noise is generated by randomly replacing the head or tail entity of a correct triplet, which is used in previous works \cite{kael}. 
Semantically-similar noise uses entities that have co-occurrence with their relations, which means that it is semantically related to the relation.
We test two typical models CAGED \cite{caged} and TransE \cite{TransE} on the \fb \cite{FB15K} and \wn \cite{WN18RR} datasets to verify whether existing works can deal with more realistic noise.
As shown in Table \ref{tab:empirical}, we add 5\% of the two types of noise to the datasets and show the precision of the top-1\% and top-5\% detected triplets.
The results show that, although the existing methods CAGED and TransE perform well on random noise, their effects are greatly reduced on semantically-similar noise. 
This is because existing models predominantly rely on graph structure, ignoring the rich textual information of KGs.
The incompleteness of KGs leads to the lack of important information that can distinguish semantically-similar noise, which exhibits greater semantic relevance and shares a similar graph structure with correct triplets.

Furthermore, we consider the prevalent scenarios of KG error detection to build adversarial noise.
For automatic KG construction and completion, noise is inevitably introduced.
An effective error detection model should possess the capability to pinpoint triplets that have been inaccurately completed or constructed.
We filter out the error results from the construction model to constitute adversarial noise.
Subsequent experiments also show that textual information plays a key role in the noise generated by completion which relies on graph structure.

Existing works need a potent method to discern more realistic noise and extract the full potential of textual information within KGs for error detection.
To achieve this goal, we opt to leverage a pre-trained language model (PLM) as the encoder of textual information in our model.
We refer to KG-BERT \cite{KGbert}, PKGC \cite{PKGC}, MLMLM \cite{MLMLM}, and CoLE \cite{CoLE}, which perform well in KG embedding and link prediction.
A PLM uses a large number of open-domain corpora for training and can supplement rich information for triplets.

We propose a novel KG error detection model \modelname.
It leverages the reconstruction of triplets to comprehend noise patterns from both textual and graph structural perspectives.
Also, we design interactive contrastive learning to align the latent representations of textual and structural information.
It facilitates noise identification based on disparities between these two forms of information.
\modelname combines the reconstruction and contrastive learning output and generates pseudo-labels to represent triplet confidence.
This adaptive confidence guides model training by alleviating noise interference and transferring knowledge between reconstruction and contrastive learning.
With the utilization of textual information, \modelname not only outperforms the state-of-the-art methods on random noise but also performs more prominently in semantically-similar noise and adversarial noise, validating the effectiveness of \modelname in complex real-world scenarios.

In summary, this paper makes the following contributions:
\begin{itemize}
\item We propose an end-to-end KG error detection model, which fully leverages both textual and structural information by reconstructing triplets, and alleviates the interference of noise. 
It transfers the knowledge between reconstruction and contrastive learning.

\item We design interactive contrastive learning to align the latent representations of textual and structural information.
We use different negative sampling policies to mine anomalous features on the alignment of latent spaces.

\item We construct two kinds of noise, semantically-similar noise and adversarial noise, to evaluate the performance of our model in more realistic scenarios.
Experiments show that \modelname not only surpasses the state-of-the-art competitors on random noise but also achieves more significant results on the datasets with semantically-similar noise and adversarial noise.
Datasets and source code are available at \url{https://github.com/nju-websoft/CCA}.
\end{itemize}

%====================%
\section{Related Work}

\subsection{KG Error Detection}
%Many studies are dedicated to exploring error detection models on KGs, which can be categorized into three groups:

The embedding-based models, e.g., TransE \cite{TransE}, DistMult \cite{DistMult}, and ComplEx \cite{Complex}, evaluate triplets by designing score functions according to the representations of entities and relations. 
TransE employs $\lVert \mathbf{h} + \mathbf{r} - \mathbf{t} \rVert$ as the score function to assess triplets. 
Subsequent works such as CKRL \cite{CKRL}, CAGED \cite{caged}, and SCEF \cite{scef} adopt the TransE's score function as a component. 
The embedding-based models heavily rely on negative sampling, and their effectiveness is hampered by the challenge of accurately modeling the real noise distribution. 
Consequently, the negative samples often deviate from the real noise, undermining the error detection performance. 

Some works exploit path information in KGs as a reference. 
PTransE \cite{PTransE} integrates relation path information into triplet embedding learning. 
KGTtm \cite{KGTtm} leverages path information for error detection.

Several works recognize that relying solely on graph structure may not be sufficient for robust KG error detection. 
Integrating additional information can offer more semantic insights into triplets or provide more accurate supervision signals, thereby enhancing model capabilities.
Defacto \cite{defacto} uses the information from relevant web pages to verify triplets, while the categorical information of entities and relations is also useful to assess the trustworthiness of triplets \cite{statistics}. 
CrossVal \cite{CrossVal} introduces human-curated knowledge repositories to assess triplet confidence in the target KG through the correlation of triplets in two KGs.
Crowdsourcing and active learning models, such as KAEL \cite{kael}, show promising results.
CKRL \cite{CKRL} and PGE \cite{PGE} also explore confidence-aware methods. 
Their confidence constraints are based solely on the score function of the TransE's loss.

The existing models mainly focus on utilizing graph structure while under-utilizing valuable textual information.
By considering this, our proposed model \modelname seeks to overcome these limitations by jointly leveraging textual and graph structural information, aiming to achieve more accurate and comprehensive KG error detection.

%--------------------%
\subsection{PLMs for KGs}
While KG error detection using PLMs is relatively scarce, the use of PLMs has demonstrated success in various tasks like KG completion. 
KG-BERT \cite{KGbert} and PKGC \cite{PKGC} achieve good results by directly fine-tuning PLMs by the triplet classification tasks using carefully constructed input templates. 
MMLLM \cite{MLMLM} effectively employs the [MASK] token in prompts to enhance link prediction.

To fuse language models with structural models, \cite{text_inhencement1,text_inhencement2,csprom-kg} leverage the text enhancement method to represent entities by combining textual and structural features.
CoLE \cite{CoLE} explores the use of the two types of information to construct models independently and uses knowledge distillation to complement and improve each other's performance.

%====================%
\section{The Proposed Model}

In this section, we describe the proposed model \modelname in detail.
The framework is shown in Figure~\ref{fig:picture001}.
The goal of the proposed model is to better leverage both textual and graph structural information at the same time. 
Given a triplet $(h, r, t)$, we first construct its input sequence by its subgraph and the descriptions of the entities and the relation.
Then, BERT and a Transformer-based encoder are used to extract textual features and graph structural features, respectively.
Reconstruction loss is employed to construct the head or tail entity in the triplet and evaluate the triplet's trustworthiness. 
Interactive contrastive learning utilizes the projection of textual and structural representations to discover anomalous features in the two alignment spaces.
The scores from contrastive learning and reconstruction classifier are aggregated to generate a pseudo label as the confidence to dynamically constrain the training process.

%--------------------%
\subsection{Feature Extraction}

\noindent\textbf{Text encoder.}
Currently PLMs like BERT \cite{bert} have been widely used to extract textual features from KGs.
For each entity $e$, following the previous work \cite{CoLE}, we leverage its description $D_e$ and human-readable literal name $N_e$ to learn the textual representation.
We first add the corresponding representation of the new entity into the PLM's vocabulary.
To better leverage the description of the entity, pre-trained prompts are constructed for initializing the new token embedding $\mathbf{e}$:
\begin{align}
\mathbf{e} = \mathrm{BERT}(\text{``The description of [MASK] is $D_e$''}),
\end{align}
where [MASK] denotes the missing entity and we use its embedding to initialize the token embedding $\mathbf{e}$.
After pre-training on the description, we construct two prompts through the masked head and tail entities to better leverage textual information.

Given a triplet $(h,r,t)$, we use the prompts presented in Table \ref{tab:prompts} as the input sequence.
In $P_h^T$, the [MASK] token is used to replace the original head entity, and the tail entity $t$ in $P_t^T$ is also masked, so that $h$ and $t$ can be reconstructed respectively by the encoder.
[CLS] and [SEP] are two special tokens used to separate the different parts of the prompt. 
[SEP]$_i^r$ is the $i$-th adjustable soft prompt token for the relation $r$, as a more expressive separator.
We provide the textual description $D_t$ of the tail entity in $P_h^T$, and $D_h$ of the head entity is used in $P_t^T$.
If the tail entity $t$ is an anomalous entity, it is hard to reconstruct $t$ through the text description of $h$ in $P_t^T$, so the reconstruction loss can be used to evaluate the error possibility of triplets from the textual view.

\begin{table}
\centering
\small
{
\begin{tabular}{l}
\hline
 $P_t^T$\,=\,[CLS]\,[SEP]$_1^r$\,$N_h$\,[SEP]$_2^r$\,$N_r$\,[SEP]$_3^r$\,[MASK]\,[SEP]$_4^r$\,$D_h$ \\
 $P_h^T$\,=\,[CLS]\,[SEP]$_1^r$\,[MASK]\,[SEP]$_2^r$\,$N_r$\,[SEP]$_3^r$\,$N_t$\,[SEP]$_4^r$\,$D_t$  \\
\hline

 $P_t^S$\,=\,[CLS]\,[$h$]\,[SEP]\,[$r$]\,[SEP]\,[MASK]\,[SEP]\,[$h'$]\,[$r'$]...\\
$P_h^S$\,=\,[CLS]\,[MASK]\,[SEP]\,[$r$]\,[SEP]\,[$t$]\,[SEP]\,[$t'$]\,[$r'$]...\\
\hline
\end{tabular}}
\caption{Inputs used in the textual and structural encoders}
\label{tab:prompts}
\end{table}

\noindent\textbf{Structure encoder.}
To make full use of the graph structure in KGs, we build an entity's subgraph by random sampling and use Transformer to encode it. 
Given a triplet $(h,r,t)$, we also create two input sequences $P_h^S$ and $P_t^S$ by masking $h$ and $t$, respectively.
For $P_h^S$, let $InNeighbor(h) = \big\{(h', r')\,|\,(h', r', h)\in\Gamma\big\}$ and $OutNeighbor(h) = \big\{(t',r')\,|\,(h, r', t')\in\Gamma\big\}$ denote two neighbor sets of entity $h$, where $\Gamma$ is the triplet set.
We randomly select some neighbors and separate them by the [SEP] token, as shown in Table~\ref{tab:prompts}. 
$P_t^S$ is constructed in the same way.

The structure encoder can learn the differences between $t$ and the subgraph of $h$.  
Therefore, by reconstructing entities from the subgraphs, errors can be distinguished.

%--------------------%
\subsection{Reconstruction Classifier}
Given a triplet $(h, r, t)$, we can get the output representation $\mathbf{e}_h^T$ of the masked head entity from the text encoder:
\begin{align}
\label{eq:get_mask_embedding}
\mathbf{e}_h^T = \mathrm{BERT}(P_h^T),
\end{align}
where $P_h^T$ is the input sequence that we construct in Table~\ref{tab:prompts}. 

Then, we can calculate the prediction logits against all candidate entities and get the cross-entropy loss $\mathcal{L}_h^T$ according to the corresponding labels:
\begin{align}
\label{eq:crossentropy}
\mathcal{L}_h^T = CE\Big(\mathrm{softmax}\big(\mathrm{MLP}(\mathbf{e}_h^T), \mathbf{E}^T\big), \mathbf{y}_h\Big),
\end{align}
where $\mathbf{E}^T$ is the matrix of all entity embeddings and $\mathbf{y}_h$ is the corresponding labels.
$\mathbf{e}_h^T$ is transferred by a multi-layer perceptron and used to calculate the prediction logits with all embeddings in $\mathbf{E}^T$. 
$CE()$ denotes the cross-entropy loss.
Note that the tail entity should also be reconstructed, and we calculate $\mathbf{e}_t^T$ and $\mathcal{L}_t^T$ in the same way.

Next, we introduce adaptive confidence to the reconstruction loss and obtain the final text reconstruction loss:
\begin{align}
\label{eq:structure_loss}
\mathcal{L}_\mathrm{text} = \sum_{(h, r, t) \in \Gamma} c(h,r,t)\big(\mathcal{L}_h^T + \mathcal{L}_t^T\big), 
\end{align}
where the triplet confidence $c(h,r,t)$ is calculated in Section~\ref{knowledgefusion} shortly. 
Note that we use the same way on the Transformer encoder to get the masked entity losses $\mathcal{L}_h^S$ and $\mathcal{L}_t^S$, and the overall reconstruction loss $\mathcal{L}_{struct}$ similar to Eqs.~(\ref{eq:get_mask_embedding}), (\ref{eq:crossentropy}), and (\ref{eq:structure_loss}).
We add $\mathcal{L}_{text}$ and $\mathcal{L}_{struct}$ and jointly train the final reconstruction loss $\mathcal{L}_\mathrm{reconstruct}$ as follows:
\begin{align}
\label{reconstruction_loss}
\mathcal{L}_\mathrm{reconstruct} = \alpha \mathcal{L}_\mathrm{text} + (1-\alpha) \mathcal{L}_\mathrm{struct}, 
\end{align}
where $\alpha$ is a hyperparameter to balance the training process.

% Finally, we use $CE()$ as the score function $score_\mathrm{text}$ and $score_\mathrm{struct}$:
Finally, we use the cross-entropy loss to compute the scores of the triplet $(h, r, t)$:
\begin{align}
score_\mathrm{text} &= \mathcal{L}_h^T + \mathcal{L}_t^T, \\
score_\mathrm{struct} &= \mathcal{L}_h^S + \mathcal{L}_t^S,
\end{align}
where $score_\mathrm{text}$ and $score_\mathrm{struct}$ are the confidence scores from text and structure reconstruction, respectively. 

%--------------------%
\subsection{Interactive Contrastive Learning}
We use interactive contrastive learning \cite{MCL} to learn the differences between textual and structural information. 
For a triplet $(h,r,t)$, $\mathbf{v}_i^h$ and $\mathbf{v}_i^t$ represent the token embeddings generated from the PLM by $P_h^T$ and $P_t^T$, respectively.
$\mathbf{u}_i^h$ and $\mathbf{u}_i^t$ represent the embeddings from the structure encoder.
$\mathbf{v}_i^h$ contains the textual information of $t$ and the reconstruction of $h$ from the PLM. 
$\mathbf{u}_i^t$ has the structural information of $h$ and the reconstruction of $t$ from Transformer.
Intuitively, $\mathbf{v}_i^h$ and $\mathbf{u}_i^t$ should be aligned, so that the prediction of $h$ can match its structural information and the prediction of $t$ can match its textual information.
Therefore, we use $\mathbf{v}_i^h$ and $\mathbf{u}_i^t$ as one anchor pair and $\mathbf{v}_i^t$ and $\mathbf{u}_i^h$ as the other pair for contrastive learning.

\noindent\textbf{Negative sampling.}
In interactive contrastive learning, we maximize the agreement between the anchor pair like $\mathbf{v}_i^h$ from the textual encoder and its structural representation $\mathbf{u}_i^t$ for one triplet $(h,r,t)$. 
We use two negative sampling policies to support stable training.

First, we construct negative samples to align the latent spaces from the two encoders. 
For the anchor sample $\mathbf{v}_i^h$ from the text encoder, another sample in the structure encoder like $\mathbf{u}_j^t$ should keep a distance from the anchor sample.
Thus, we randomly take another sample from the structure encoder as one of the negative samples.

To improve the sensitivity and robustness of the model against noise, we construct error samples as negative examples based on the anchor.
We randomly replace the head (or tail) entity of the anchor sample $(h,r,t)$ to construct a typical error sample $(h,r,t')$ (or $(h',r,t)$), and generate its representations $\mathbf{m}_{i_k}^h$ and $\mathbf{n}_{i_k}^t$ from different encoders, respectively.
Note that $(h,r,t')$ is in fact the noise generated from disturbance. 
Therefore, $\mathbf{m}_{i_k}^h$ and $\mathbf{n}_{i_k}^t$ should not be aligned.

To reduce the training cost, we directly use the representation of the entity obtained by the encoder in each training batch to disturb original embeddings.
For a triplet $(h, r, t)$, the similarity of its negative samples is calculated as
\begin{align}
\begin{split}
neg\_sim(h, r, t) &= \sum\limits_{j\neq i} \exp\big(sim(\mathbf{v}_i^h, \mathbf{u}_j^t)\big) \\
&\quad + \sum\limits_{x =1}^{X} \exp\big(sim(\mathbf{m}_{i_x}^h, \mathbf{n}_{i_x}^t)\big),
\end{split}
\end{align}
where $\mathbf{v}_i^h$ denotes the masked head triplet embedding of the $i$-th sample and $\mathbf{u}_j^t$ denotes the masked tail triplet embedding of the $j$-th sample.
$sim()$ denotes the cosine similarity used to calculate the distance of the two embeddings.
$\mathbf{m}_{i_x}^h$ and $\mathbf{n}_{i_x}^t$ represent the randomly replaced embeddings from $\mathbf{v}_i^h$ and $\mathbf{u}_i^t$, and we replace them for $X=4$ times randomly.

\noindent\textbf{Adaptive contrastive learning.}
We employ the InfoNCE loss \cite{InfoNCE} to train interactive contrastive learning. For the anchor pair $\mathbf{v}_i^h$ and $\mathbf{u}_i^t$, the loss $\mathcal{L}_\mathrm{ICL}^1(h,r,t)$ is
\begin{align}
\label{eq:ICL_loss}
\resizebox{\columnwidth}{!}{$
\mathcal{L}_\mathrm{ICL}^1(h,r,t) = - \log \frac{ \exp(sim(\mathbf{v}_i^h, \mathbf{u}_i^t))}{ \exp(sim(\mathbf{v}_i^h, \mathbf{u}_i^t)) + neg\_sim(h, r, t)},
$}
\end{align}
where $neg\_sim(h, r, t)$ is the similarity of negative samples.
Note that if we choose $\mathbf{v}_i^t$ and $\mathbf{u}_i^h$ as the anchor pair, we can get $\mathcal{L}_\mathrm{ICL}^2(h,r,t)$ in the same way.
Therefore, the final contrastive training loss is
\begin{align}
\label{eq:contrastive_loss}
\resizebox{\columnwidth}{!}{$
\mathcal{L}_\mathrm{contr} = \sum\limits_{(h,r,t)\in \Gamma}c(h, r, t)\big(\mathcal{L}_\mathrm{ICL}^1(h,r,t) + \mathcal{L}_\mathrm{ICL}^2(h,r,t)\big),
$}
\end{align}
where $c(h,r,t)$ is the adaptive confidence of $(h, r, t)$. 
The score function of contrastive learning of $(h,r,t)$ is
\begin{align}
\label{eq:contrastive_loss_score}
score_\mathrm{contrastive} = sim(\mathbf{v}_i^h, \mathbf{u}_j^t).
\end{align}

%--------------------%
\subsection{Knowledge Fusion}
\label{knowledgefusion}

Since $score_\mathrm{text}$ and $score_\mathrm{struct}$ are both logits from triplet reconstruction, we directly add them to obtain the reconstruction score:
\begin{align}
score_\mathrm{reconstruct} = score_\mathrm{text} + \lambda\,score_\mathrm{struct},
\end{align}
where $\lambda$ is the hyperparameter to balance the scores.
Note that the scores from reconstruction and contrastive learning have quite different distributions, so we use ranking to combine them.
By ranking the error scores $score_\mathrm{reconstruct}$ and $score_\mathrm{contrastive}$, we obtain the error ranks $R_1(h,r,t)$ and $R_2(h,r,t)$ of each triplet, respectively.
The final score is
\begin{align}
\label{eq:final_loss_score}
score(h,r,t)=\frac{1}{\lceil\frac{R_1(h,r,t)}{\gamma}\rceil}+\frac{1}{\lceil\frac{R_2(h,r,t)}{\gamma}\rceil ^\beta},
\end{align}
where $\gamma$ is the hyperparameter to control the number of triplets with the same score. 
$\beta$ is used to balance the scores.

We generate a pseudo label from the final score.
Let $Z = normalize\big([z_1,\dots,z_n]\big)$ be the pseudo label set, where $z_i\sim N(\mu,\rho)$ and $Z$ is sorted in ascending order, the adaptive confidence of $(h,r,t)$ is as follows:
\begin{align}
c(h, r, t) = z_{R(h, r, t)},
\end{align}
where $R(h,r,t)$ is the rank of $score(h,r,t)$ and we use $c(h, r, t)$ as the final result.
Note that $c(h, r, t)$ is the result integrated from textual and structural reconstruction and contrastive learning.
We use $c(h, r, t)$ directly constrain the training process according to Eqs.~(\ref{eq:structure_loss}) and (\ref{eq:contrastive_loss}) to transfer knowledge across components.

%====================%
\section{Experiments and Results}
%In this section, we report the experimental results of the proposed model.

\subsection{Dataset Construction} 
We conduct our experiments on \fb \cite{FB15K} and \wn \cite{WN18RR}.
%They are sampled from Freebase and WordNet, respectively.
We add 5\% noise into these two datasets. 
The statistics of the new datasets are shown in Table~\ref{tab:dataset}.
We construct three types of noise to better evaluate the performance of our model:

\begin{table}
\centering
\small
{
\begin{tabular}{l|ccc}
\toprule
Datasets & Correct triplets & Wrong triplets & Avg. deg.\\
\midrule 
\fb & 310,116 & 16,321 & 44.89\\
\wn & \ \ 93,003 & \ \ 4,894 & \ \ 4.78\\
\bottomrule
\end{tabular}}
\caption{Statistics of the modified datasets}
\label{tab:dataset}
\end{table}

\begin{itemize}
\item \textbf{Random noise}, drawn from a uniform distribution that does not depend on the data and is not predictable.
For a correct triplet $(h,r,t)$, random noise is constructed by randomly replacing one of the entities or the relation.

\item \textbf{Semantically-similar noise}, which is more realistic because errors primarily stem from semantic confusion in real-world scenarios.
To better evaluate semantically-related error detection, we introduce semantically-similar noise.
Given a triplet $(h,r,t)$, we form a candidate set $S_t$ by selecting other tail entities linked to $r$. 
For each entity $e \in S_t$, we employ BERT to encode its description, acquiring the semantic embedding $\mathbf{e}$ of $e$. 
We calculate the sampling probability distribution of semantic similarity as follows and utilize it as a sampling probability to replace the original $t$:
\begin{align}
\label{eq:distribution}
\mathbf{P}(t) = \mathrm{softmax}\Big(\big[\mathbf{t} \cdot \mathbf{e}_1, \mathbf{t} \cdot \mathbf{e}_2,\dots, \mathbf{t} \cdot\mathbf{e}_i\big]\Big),
\end{align}
where $\mathbf{t}$ denotes the embedding of the original tail entity $t$, $\mathbf{e}_i$ denotes the embedding of the $i$-th candidate entity, and $\cdot$ denotes the dot product.

\item \textbf{Adversarial noise}, which is adversarially generated from KG construction models.
Since error detection models are frequently employed to discern errors in automatically constructed KGs, we use TransE \cite{TransE} for adversarial noise generation and evaluate the ability of models to recognize errors during KG construction.
Given a dataset $D$, we randomly divide it into training and testing sets, $D_{train}$ and $D_{test}$. 
After training on $D_{train}$, we randomly select one entity within the top-10 prediction results of the triplet in $D_{test}$ to construct its error triplet. 
We iteratively repeat this process until we get an adequate amount of noise.
\end{itemize}

\begin{table*}
\small
{
\begin{tabular}{c|c|c|ccccc|ccccc}
\toprule
& \multirow{2}{*}{\makecell[c]{Model\\types}} & \multirow{2}{*}{Models} & \multicolumn{5}{c|}{\fb} & \multicolumn{5}{c}{\wn} \\
\cmidrule(lr){4-8} \cmidrule(lr){9-13} 
& & & $K$=1\% & $K$=2\% & $K$=3\% & $K$=4\% & $K$=5\% & $K$=1\% & $K$=2\% & $K$=3\% & $K$=4\% & $K$=5\% \\ 
\midrule
\multirow{9}{*}{\rotatebox{90}{Precision@top-$K$}} 
& \multirow{5}{*}{\makecell[c]{Struct\\models}} 
& TransE        & 0.946 & 0.774 & 0.606 & 0.498 & 0.423 & 0.690 & 0.576 & 0.501 & 0.437 & 0.400\\
& & DistMult    & 0.764 & 0.630 & 0.530 & 0.463 & 0.410 & 0.687 & 0.633 & 0.526 & 0.438 & 0.374\\
& & ComplEx     & 0.802 & 0.633 & 0.521 & 0.446 & 0.393 & 0.774 & 0.699 & 0.550 & 0.449 & 0.384\\

&  & KGTtm       & 0.857 & 0.687 & 0.631 & 0.467 & 0.437 & 0.789 & 0.644 & 0.541 & 0.473 & 0.417\\ 
& & CAGED       & 0.863 & 0.666 & 0.602 & 0.543 & 0.467 & 0.753 & 0.620 & 0.536 & 0.470 & 0.421\\
\cmidrule(lr){2-13} & \multirow{4}{*}{\makecell[c]{Text\\models}} 
 & KG-BERT     & 0.966 & 0.799 & 0.660 & \underline{0.584} & 0.498 & 0.973 & \textbf{0.968} & \textbf{0.938} & \underline{0.829} & \underline{0.710}\\
&  & StAR     & \textbf{0.970} & \textbf{0.835} & 0.681 & 0.571 & 0.490 & 0.971 & 0.918 & 0.842 & 0.739 & 0.647\\
&   & CSProm-KG     & 0.961 & 0.798 & \underline{0.689} & 0.574 & \underline{0.509} & \underline{0.977} & 0.927 & 0.869 & 0.773 & 0.680\\
& & \modelname (ours)        & \underline{0.969} & \underline{0.812} & \textbf{0.707} & \textbf{0.599} & \textbf{0.534} & \textbf{0.986} & \underline{0.959} & \underline{0.920} & \textbf{0.834} & \textbf{0.733}\\   
\midrule

\multirow{9}{*}{\rotatebox{90}{Recall@top-$K$}} 
& \multirow{5}{*}{\makecell[c]{Struct\\models}} 
& TransE        & 0.189 & 0.310 & 0.364 & 0.399 & 0.423 & 0.138 & 0.231 & 0.300 & 0.350 & 0.400 \\
& & DistMult    & 0.153 & 0.252 & 0.318 & 0.371 & 0.410 & 0.137 & 0.253 & 0.316 & 0.350 & 0.374 \\
& & ComplEx     & 0.161 & 0.254 & 0.313 & 0.357 & 0.393 & 0.155 & 0.279 & 0.330 & 0.359 & 0.384 \\

&  & KGTtm       & 0.171 & 0.275 & 0.378 & 0.374 & 0.437 & 0.158 & 0.257 & 0.324 & 0.378 & 0.417 \\ 
& & CAGED       & 0.173 & 0.266 & 0.362 & 0.435 & 0.467 & 0.150 & 0.248 & 0.321 & 0.376 & 0.421 \\ 
\cmidrule(lr){2-13} & \multirow{4}{*}{\makecell[c]{Text\\models}} \
& KG-BERT     & \underline{0.193} & 0.319 & 0.396 & \underline{0.467} & 0.498 & \underline{0.195} & \textbf{0.387} & \textbf{0.563} & \underline{0.663} & \underline{0.710} \\
& & StAR   & \textbf{0.194} & \textbf{0.334} & 0.409 & 0.457 & 0.490 & 0.194 & 0.367 & 0.505 & 0.591 & 0.647 \\  
& & CSProm-KG  & 0.192 & 0.319 &\underline{0.413} & 0.459 & \underline{0.509} & \underline{0.195} & 0.371 & 0.521 & 0.618 & 0.680 \\  
& & \modelname (ours)        & \textbf{0.194} & \underline{0.325} &\textbf{0.424} & \textbf{0.479} & \textbf{0.534} & \textbf{0.197} & \underline{0.384} & \underline{0.552} & \textbf{0.667} & \textbf{0.733} \\  
\bottomrule
\end{tabular}}
\caption{Results of precision and recall at top-$K$ on \fb and \wn}
\label{tab:main_results}
\end{table*}

%--------------------%
\subsection{Settings}

We use the ranking measures for evaluation. 
All triplets in a KG are ranked based on the confidence scores in ascending order. 
Triplets with lower confidence scores are more likely to be noisy.
We follow CAGED \cite{caged} and use precision@top-$K$ and recall@top-$K$ to assess the performance, where $K$ denotes the ratio (e.g., 5\%).

All experiments are conducted on two Intel Xeon Gold 6326 CPUs, 512GB RAM, and one NVIDIA RTX A6000 GPU.
We leverage the BERT-base model from huggingface as the PLM.
We use PyTorch to implement our model and employ the AdamW optimizer and a cosine decay scheduler with a linear warm-up for optimization.
The grid search is used for hyperparameter tuning.
The results of KG embedding models are obtained from $\mu$KG \cite{muKG}, a recent open-source library for KG embedding.

%--------------------%
\subsection{Baseline Models}
We compare our model with eight baselines, including five structural models and three textual models.
\begin{itemize}
\item \textbf{Structural models.} 
We choose three typical embedding models TransE \cite{TransE}, DistMult \cite{DistMult}, and ComplEx \cite{Complex} and two state-of-the-art structural error detection models KGTtm \cite{KGTtm} and CAGED \cite{caged} for comparison. Both KGTtm and CAGED utilize graph structural information based on the TransE's score function.

\item \textbf{Textual models.} 
We choose one textual classification model KG-BERT \cite{KGbert}, and two recent models StAR \cite{text_inhencement2} and CSProm-KG \cite{csprom-kg}, which leverage both textual and structural information, for comparison.
KG-BERT takes entity and relation descriptions of a triplet as input and computes scores by binary classification. StAR and CSProm-KG use the representations obtained from PLMs to aid structural models. 
CCA also falls into this category.
\end{itemize}

\begin{table}
\centering
\small
{
\begin{tabular}{c|l|ccc}
\toprule
& \multicolumn{1}{c|}{Models} & $K$=1\% & $K$=3\% & $K$=5\% \\
\midrule
\multirow{5}{*}{\rotatebox{90}{\fb}}
& \modelname (full) & .969 / .194 & .707 / .424 & .535 / .535 \\
& -- adapt conf.    & .951 / .190 & .664 / .398 & .496 / .496 \\  
& -- inter contr.   & .961 / .192 & .679 / .407 & .509 / .509 \\
& -- struct recon.  & .797 / .159 & .582 / .349 & .475 / .475 \\
& -- text recon.    & .778 / .156 & .543 / .325 & .414 / .414 \\ 
\midrule
\multirow{5}{*}{\rotatebox{90}{\wn}}
& \modelname (full) & .986 / .197 & .920 / .552 & .733 / .733 \\
& -- adapt conf.    & .971 / .194 & .858 / .515 & .632 / .632 \\  
& -- inter contr.   & .979 / .196 & .911 / .546 & .722 / .722 \\
& -- struct recon.  & .974 / .195 & .914 / .549 & .726 / .726 \\
& -- text recon.    & .577 / .115 & .513 / .308 & .423 / .423 \\ 
\bottomrule
\end{tabular}}
\caption{Ablation results of precision and recall at top-$K$}
\label{tab:ablation_study}
\end{table}

%--------------------%
\subsection{Main Results}
Table~\ref{tab:main_results} presents the comparison results of our model and eight baseline models on \fb and \wn,
where we add 5\% noise, containing three types of noise in equal quantities.
Overall, our model outperforms eight baseline models on both datasets.
We have three observations below:

First, compared with the KG embedding models, the error detection models generally perform better.
This is because the KG embedding models assume that all triplets are correct.
They learn representations of entities and relations without alleviating disturbance from noise, making it difficult to discriminate error triplets.
With adaptive confidence, \modelname can effectively improve performance.

Second, benefiting from PLMs, the textual models outperform the structural models on the two datasets.
They both can learn noise patterns in the textual view with the descriptions of entities and relations, considering that PLMs can capture factual knowledge from a large amount of open-domain corpora. 
\modelname combines both textual information and structural information and uses a Transformer encoder to learn structural information from scratch, which is more effective than other textual models.

\begin{table}[!t]
\centering
\small
{
\begin{tabular}{c|l|ccc}
\toprule
& Models & Random & Similar & Adversarial \\
\midrule
\multirow{9}{*}{\rotatebox{90}{\fb}}
& TransE      & 0.726 & 0.304 & 0.125 \\
& ComplEx     & 0.734 & 0.306 & 0.150 \\ 
& DistMult    & 0.662 & 0.328 & 0.125 \\
& KGTtm       & 0.730 & 0.318 & 0.128 \\ 
& CAGED       & \underline{0.758} & 0.331 & 0.126 \\ 
& KG-BERT     & 0.674 & 0.336 & 0.199 \\ 
& StAR        & 0.728 & 0.371 & 0.164 \\
& CSProm-KG   & 0.732 & \underline{0.419} & \underline{0.211} \\
& \modelname (ours)        & \textbf{0.768} & \textbf{0.453} & \textbf{0.240} \\
\midrule
\multirow{9}{*}{\rotatebox{90}{\wn}}
& TransE      & 0.434 & 0.351 & 0.314 \\
& ComplEx     & 0.384 & 0.303 & 0.366 \\ 
& DistMult    & 0.316 & 0.338 & 0.285 \\
& KGTtm       & 0.448 & 0.391 & 0.359 \\ 
& CAGED       & 0.486 & 0.388 & 0.373 \\ 
& KG-BERT     & \underline{0.806} & 0.633 & \textbf{0.599} \\
& StAR        & 0.794 & 0.610 & 0.527 \\
& CSProm-KG   & 0.791 & \underline{0.634} & 0.536 \\
& \modelname (ours)        & \textbf{0.807} & \textbf{0.657} & \underline{0.545} \\  
\bottomrule
\end{tabular}}
\caption{Precision at top-5\% on three different error types}
\label{tab:noise_type}
\end{table}

\begin{table*}
\centering
\small
{
    \begin{tabular}{l|cl|ccc} 
    \toprule
        & & & \modelname & KG-BERT & CAGED \\
        \midrule
        \multirow{2}{*}{Case 1.} & Noise & Razzie Award for Worst Actor, award winner, \underline{\rm Richard D. Zanuck}   & \multirow{2}{*}{2.9\%} & \multirow{2}{*}{\ \ 7.6\%} & \multirow{2}{*}{\ \ 3.4\%}\\
        & Truth & Razzie Award for Worst Actor, award winner, {\rm Marlon Wayans} \\
        \midrule
        \multirow{2}{*}{Case 2.} & Noise & \underline{Tony Award for Best Choreography}, ceremony, 54th Academy Awards & \multirow{2}{*}{3.2\%} & \multirow{2}{*}{10.6\%}   & \multirow{2}{*}{11.6\%}\\
        & Truth & Academy Award for Best Sound Mixing, ceremony, 54th Academy Awards \\
        \midrule
        \multirow{2}{*}{Case 3.} & Noise  & \underline{Lauren Tom}, place of birth, Buenos Aires & \multirow{2}{*}{2.2\%} & \multirow{2}{*}{\ \ 3.2\%} & \multirow{2}{*}{35.1\%}\\
        & Truth & Sebastian Krys, place of birth,  Buenos Aires  \\
        \midrule
        %\multirow{2}{*}{Case 4.} & Noise  & \underline{The Venture Bros}, program creator, {\rm Stephen Merchant} & \multirow{2}{*}{4.6\%} & \multirow{2}{*}{7.8\%} & \multirow{2}{*}{17.7\%}\\
        %& Truth & The Office, program creator, {\rm Stephen Merchant} \\
        %\midrule
        \multirow{2}{*}{Case 4.} & Noise  & Viola Davis, award nominee, \underline{\rm Carrie-Anne Moss} & \multirow{2}{*}{1.0\%} & \multirow{2}{*}{10.9\%} & \multirow{2}{*}{\ \ 4.3\%} \\
        & Truth  &  Viola Davis, award nominee, {\rm Meryl Streep} \\      
    \bottomrule
    \end{tabular}}
\caption{Case study on \fb}
\label{tab:case_study}
\end{table*}

Third, the performance gap between \modelname and KG-BERT on \wn is less significant than that on \fb.
We think the reason is that \wn is much sparser than \fb.
As shown in Table \ref{tab:dataset}, the average degree of entities in \wn is far less than that of \fb, which indicates that there is less structural information in \wn.
Thus, our model obtains less improvement by combining with graph structure on sparse KGs.

%--------------------%
\subsection{Ablation Study}
We conduct an ablation study to assess the impact of each component in \modelname.
We have four variants of our model by removing confidence adaption, interactive contrastive learning, structure reconstruction, or text reconstruction. 
We report their performance under the same settings.

Table~\ref{tab:ablation_study} shows that all components contribute to our model on both datasets, where text reconstruction contributes the most.
On \fb, the precision improvement of text construction is from 0.414 to 0.535.
This shows that applying PLMs to KG error detection is a promising way, and it deserves further research.
The improvement is more obvious on \wn.
The reason is that the structure-based models perform poorly, as \wn is sparser than \fb.

Removing adaptive confidence affects differently on the two datasets.
On \wn, adaptive confidence contributes more than \fb, as the performance gap between text and structure reconstruction is larger.
Without adaptive confidence, structure reconstruction is hard to gain benefits from the knowledge of PLMs, which provides scores with lower accuracy and disturbs the final results.
Therefore, adaptive confidence can be more effective in improving overall performance when components have a larger performance gap.

%--------------------%
\subsection{Performance on Different Error Types}
To investigate the robustness of all models to different noise types, we add three types of 5\% noise separately to the datasets.
Note that precision@top-5\% and recall@top-5\% have the same value as we add 5\% noise.
Table~\ref{tab:noise_type} presents precision@top-5\% on \fb and \wn.

On \fb, \modelname outperforms all baselines.
For random noise, the structure-based models generally outperform KG-BERT, as this type of noise can be well distinguished by graph structure alone.
For semantically-similar noise, all baseline models perform closely.
As for adversarial noise, the textual models outperform the structural ones, because this type of noise consists of entities that are wrongly predicted as correct entities by TransE, which is a structural model.
Our \modelname takes advantage of the knowledge from PLMs and graph structure at the same time, so it outperforms all baselines, especially for semantically-similar noise.

On \wn, \modelname is comparable to KG-BERT, which is different from the observations on \fb. 
The models with PLMs largely outperform the structural models on all three types of noise, mainly due to that \wn is sparser than \fb.
For adversarial noise, our model underperforms KG-BERT.
Given that the structure-based models do not work well on \wn, Transformer in \modelname provides less reliable knowledge for the PLM, hindering it from distinguishing the noise correctly.

%--------------------%
\subsection{Case Study}
To explore how textual information and graph structure act on error detection, we perform a case study on \modelname, KG-BERT, and CAGED.
Table~\ref{tab:case_study} shows the position of the error triplet in a ranking, where a smaller percentage is better.

In the first two cases, \modelname leverages both textual and graph structural information, and outperforms KG-BERT and CAGED, which solely use one type of information.
For example, in Case 1, the description shows that Richard D. Zanuck is an American film producer, and the graph structural information records films that he has produced.
Evidence from text and graph structure are complementary to each other.
In Case 3, CAGED is not effective compared with \modelname and KG-BERT, which is caused by the lack of graph structural information.
The degrees of entities ``Lauren Tom'' and ``The Venture Bros'' are 15 and 17, respectively, which are much smaller than the average degree of 44.89 in \fb.
In Case 4, entities in the noise and correct triplets all have abundant graph structural information, so CAGED can achieve a better effect.

%====================%
\section{Conclusion}
In this paper, we propose a novel KG error detection model.
It encodes textual and graph structural information to find noise patterns.
To alleviate the disturbance of noise and integrate the knowledge from the two encoders, we design a confidence adaption model to aggregate the results and constrain the training process.
To learn the noise patterns between textual and structural information, we leverage interactive contrastive learning to align latent spaces.
We construct semantically-similar noise and adversarial noise for evaluation.
Experiments show that our model achieves good results on semantically-similar noise and adversarial noise.

\section{Acknowledgments}
This work is supported by the National Natural Science Foundation of China (No. 62272219).

%\appendix

\bibliography{aaai24}

\begin{thebibliography}{31}
\providecommand{\natexlab}[1]{#1}

\bibitem[{Bordes et~al.(2013)Bordes, Usunier, Garc{\'{i}}a{-}Dur{\'{a}}n,
  Weston, and Yakhnenko}]{TransE}
Bordes, A.; Usunier, N.; Garc{\'{i}}a{-}Dur{\'{a}}n, A.; Weston, J.; and
  Yakhnenko, O. 2013.
\newblock Translating embeddings for modeling multi-relational data.
\newblock In \emph{NIPS}, 2787--2795.

\bibitem[{Carlson et~al.(2010)Carlson, Betteridge, Kisiel, Settles, Hruschka,
  and Mitchell}]{nell}
Carlson, A.; Betteridge, J.; Kisiel, B.; Settles, B.; Hruschka, E.; and
  Mitchell, T. 2010.
\newblock Toward an architecture for never-ending language learning.
\newblock In \emph{AAAI}, 1306--1313.

\bibitem[{Chen et~al.(2023)Chen, Wang, Sun, Li, and Lam}]{csprom-kg}
Chen, C.; Wang, Y.; Sun, A.; Li, B.; and Lam, K.-Y. 2023.
\newblock Dipping PLMs sauce: Bridging structure and text for effective
  knowledge graph completion via conditional soft prompting.
\newblock In \emph{Findings of ACL}, 11489--11503.

\bibitem[{Chen et~al.(2021)Chen, Liu, Gao, Jiao, Zhang, and Ji}]{hitter}
Chen, S.; Liu, X.; Gao, J.; Jiao, J.; Zhang, R.; and Ji, Y. 2021.
\newblock HittER: Hierarchical transformers for knowledge graph embeddings.
\newblock In \emph{EMNLP}, 10395--10407.

\bibitem[{Cheng et~al.(2022)Cheng, Li, Xu, Dong, and Sun}]{PGE}
Cheng, K.; Li, X.; Xu, Y.~E.; Dong, X.~L.; and Sun, Y. 2022.
\newblock PGE: Robust product graph embedding learning for error detection.
\newblock \emph{Proc. VLDB Endow.}, 15(6): 1288--1296.

\bibitem[{Clouatre et~al.(2021)Clouatre, Trempe, Zouaq, and Chandar}]{MLMLM}
Clouatre, L.; Trempe, P.; Zouaq, A.; and Chandar, S. 2021.
\newblock MLMLM: Link prediction with mean likelihood masked language model.
\newblock In \emph{Findings of ACL}, 4321--4331.

\bibitem[{Dettmers et~al.(2018)Dettmers, Minervini, Stenetorp, and
  Riedel}]{WN18RR}
Dettmers, T.; Minervini, P.; Stenetorp, P.; and Riedel, S. 2018.
\newblock Convolutional 2D knowledge graph embeddings.
\newblock In \emph{AAAI}, 1811--1818.

\bibitem[{Dong et~al.(2023)Dong, Zhang, Huang, Tan, Zha, and Zihao}]{kael}
Dong, J.; Zhang, Q.; Huang, X.; Tan, Q.; Zha, D.; and Zihao, Z. 2023.
\newblock Active ensemble learning for knowledge graph error detection.
\newblock In \emph{WSDM}, 877--885.

\bibitem[{Dong et~al.(2014)Dong, Gabrilovich, Heitz, Horn, Lao, Murphy,
  Strohmann, Sun, and Zhang}]{knowledge_vault}
Dong, X.; Gabrilovich, E.; Heitz, G.; Horn, W.; Lao, N.; Murphy, K.; Strohmann,
  T.; Sun, S.; and Zhang, W. 2014.
\newblock Knowledge vault: A web-scale approach to probabilistic knowledge
  fusion.
\newblock In \emph{KDD}, 601--610.

\bibitem[{Guo et~al.(2022)Guo, Zhuang, Qin, Zhu, Xie, Xiong, and He}]{KGRec}
Guo, Q.; Zhuang, F.; Qin, C.; Zhu, H.; Xie, X.; Xiong, H.; and He, Q. 2022.
\newblock A survey on knowledge graph-based recommender systems.
\newblock \emph{IEEE Trans. Knowl. Data Eng.}, 34(8): 3549--3568.

\bibitem[{Jia et~al.(2019)Jia, Xiang, Chen, and Wang}]{KGTtm}
Jia, S.; Xiang, Y.; Chen, X.; and Wang, K. 2019.
\newblock Triple trustworthiness measurement for knowledge graph.
\newblock In \emph{WWW}, 2865--2871.

\bibitem[{Kenton and Toutanova(2019)}]{bert}
Kenton, J. D. M.-W.~C.; and Toutanova, L.~K. 2019.
\newblock BERT: Pre-training of deep bidirectional transformers for language
  understanding.
\newblock In \emph{NAACL}, 4171--4186.

\bibitem[{Lehmann et~al.(2012)Lehmann, Gerber, Morsey, and Ngomo}]{defacto}
Lehmann, J.; Gerber, D.; Morsey, M.; and Ngomo, A.-C.~N. 2012.
\newblock Defacto-deep fact validation.
\newblock In \emph{ISWC}, 312--327.

\bibitem[{Lin et~al.(2015)Lin, Liu, Luan, Sun, Rao, and Liu}]{PTransE}
Lin, Y.; Liu, Z.; Luan, H.; Sun, M.; Rao, S.; and Liu, S. 2015.
\newblock Modeling relation paths for representation learning of knowledge
  bases.
\newblock In \emph{EMNLP}, 705--714.

\bibitem[{Liu et~al.(2022)Liu, Sun, Li, and Hu}]{CoLE}
Liu, Y.; Sun, Z.; Li, G.; and Hu, W. 2022.
\newblock I know what you do not know: Knowledge graph embedding via
  co-distillation learning.
\newblock In \emph{CIKM}, 1329--1338.

\bibitem[{Luo, Sun, and Hu(2022)}]{muKG}
Luo, X.; Sun, Z.; and Hu, W. 2022.
\newblock $\mu$KG: A library for multi-source knowledge graph embeddings and
  applications.
\newblock In \emph{ISWC}, 610--627.

\bibitem[{Lv et~al.(2022)Lv, Lin, Cao, Hou, Li, Liu, Li, and Zhou}]{PKGC}
Lv, X.; Lin, Y.; Cao, Y.; Hou, L.; Li, J.; Liu, Z.; Li, P.; and Zhou, J. 2022.
\newblock Do pre-trained models benefit knowledge graph completion? A reliable
  evaluation and a reasonable approach.
\newblock In \emph{Findings of ACL}, 3570--3581.

\bibitem[{Nadkarni et~al.(2021)Nadkarni, Wadden, Beltagy, Smith, Hajishirzi,
  and Hope}]{text_inhencement1}
Nadkarni, R.; Wadden, D.; Beltagy, I.; Smith, N.; Hajishirzi, H.; and Hope, T.
  2021.
\newblock Scientific language models for biomedical knowledge base completion:
  An empirical study.
\newblock In \emph{AKBC}.

\bibitem[{Paulheim and Bizer(2014)}]{statistics}
Paulheim, H.; and Bizer, C. 2014.
\newblock Improving the quality of linked data using statistical distributions.
\newblock \emph{Int. J. Semant. Web Inf. Syst.}, 10(2): 63--86.

\bibitem[{Saxena, Tripathi, and Talukdar(2020)}]{EmbedKGQA}
Saxena, A.; Tripathi, A.; and Talukdar, P.~P. 2020.
\newblock Improving multi-hop question answering over knowledge graphs using
  knowledge base embeddings.
\newblock In \emph{ACL}, 4498--4507.

\bibitem[{Toutanova et~al.(2015)Toutanova, Chen, Pantel, Poon, Choudhury, and
  Gamon}]{FB15K}
Toutanova, K.; Chen, D.; Pantel, P.; Poon, H.; Choudhury, P.; and Gamon, M.
  2015.
\newblock Representing text for joint embedding of text and knowledge bases.
\newblock In \emph{EMNLP}, 1499--1509.

\bibitem[{Trouillon et~al.(2016)Trouillon, Welbl, Riedel, Gaussier, and
  Bouchard}]{Complex}
Trouillon, T.; Welbl, J.; Riedel, S.; Gaussier, {\'E}.; and Bouchard, G. 2016.
\newblock Complex embeddings for simple link prediction.
\newblock In \emph{ICML}, 2071--2080.

\bibitem[{van~den Oord, Li, and Vinyals(2018)}]{InfoNCE}
van~den Oord, A.; Li, Y.; and Vinyals, O. 2018.
\newblock Representation learning with contrastive predictive coding.
\newblock \emph{arXiv}, 1807.03748.

\bibitem[{Wang et~al.(2021)Wang, Shen, Long, Zhou, Wang, and
  Chang}]{text_inhencement2}
Wang, B.; Shen, T.; Long, G.; Zhou, T.; Wang, Y.; and Chang, Y. 2021.
\newblock Structure-augmented text representation learning for efficient
  knowledge graph completion.
\newblock In \emph{WWW}, 1737--1748.

\bibitem[{Wang, Ma, and Gao(2020)}]{CrossVal}
Wang, Y.; Ma, F.; and Gao, J. 2020.
\newblock Efficient knowledge graph validation via cross-graph representation
  learning.
\newblock In \emph{CIKM}, 1595--1604.

\bibitem[{Xie et~al.(2018)Xie, Liu, Lin, and Lin}]{CKRL}
Xie, R.; Liu, Z.; Lin, F.; and Lin, L. 2018.
\newblock Does William Shakespeare really write Hamlet? Knowledge
  representation learning with confidence.
\newblock In \emph{AAAI}, 4954--4961.

\bibitem[{Yang et~al.(2015)Yang, tau Yih, He, Gao, and Deng}]{DistMult}
Yang, B.; tau Yih, S.~W.; He, X.; Gao, J.; and Deng, L. 2015.
\newblock Embedding entities and relations for learning and inference in
  knowledge bases.
\newblock In \emph{ICLR}.

\bibitem[{Yang et~al.(2022)Yang, An, Cai, and Xu}]{MCL}
Yang, C.; An, Z.; Cai, L.; and Xu, Y. 2022.
\newblock Mutual contrastive learning for visual representation learning.
\newblock In \emph{AAAI}, 3045--3053.

\bibitem[{Yao, Mao, and Luo(2019)}]{KGbert}
Yao, L.; Mao, C.; and Luo, Y. 2019.
\newblock KG-BERT: BERT for knowledge graph completion.
\newblock \emph{arXiv}, 1909.03193.

\bibitem[{Zhang et~al.(2022)Zhang, Dong, Duan, Huang, Liu, and Xu}]{caged}
Zhang, Q.; Dong, J.; Duan, K.; Huang, X.; Liu, Y.; and Xu, L. 2022.
\newblock Contrastive knowledge graph error detection.
\newblock In \emph{CIKM}, 2590--2599.

\bibitem[{Zhao and Liu(2019)}]{scef}
Zhao, Y.; and Liu, J. 2019.
\newblock SCEF: A support-confidence-aware embedding framework for knowledge
  graph refinement.
\newblock \emph{arXiv}, 1902.06377.

\end{thebibliography}

\end{document}